\documentclass[10pt, a4paper, onecolumn]{article}

\usepackage[numbers]{natbib}

\usepackage[T1]{fontenc}
\usepackage{lmodern}

\usepackage{natbib}
\usepackage[utf8]{inputenc} 
\usepackage{booktabs}       
\usepackage{amsfonts}       
\usepackage{nicefrac}       
\usepackage{microtype}      
\usepackage{xcolor}         
\usepackage{graphicx}
\usepackage{amsmath,amssymb,amsfonts}
\usepackage{bm}
\usepackage{url}
\usepackage{hyperref}
\usepackage{algorithm}
\usepackage{algorithmic}
\usepackage{amsmath}
\usepackage{mathtools}
\usepackage{colortbl}

\usepackage{array,multirow,graphicx}

\usepackage{geometry}
\newgeometry{vmargin={15mm}, hmargin={12mm,17mm}} 

\title{An Initialization Schema for Neuronal Networks on Tabular Data}

\author{
	Wolfgang Fuhl\\
	University Tübingen\\
	Tübingen, 72076 \\
	\texttt{wolfgang.fuhl@uni-tuebingen.de} \\
}

\begin{document}
	
	\maketitle
	
	\begin{abstract}
		Nowadays, many modern applications require heterogeneous tabular data, which is still a challenging task in terms of regression and classification. Many approaches have been proposed to adapt neural networks for this task, but still, boosting and bagging of decision trees are the best-performing methods for this task.
		In this paper, we show that a binomial initialized neural network can be used effectively on tabular data. The proposed approach shows a simple but effective approach for initializing the first hidden layer in neural networks. We also show that this initializing schema can be used to jointly train ensembles by adding gradient masking to batch entries and using the binomial initialization for the last layer in a neural network. For this purpose, we modified the hinge binary loss and the soft max loss to make them applicable for joint ensemble training. 
		We evaluate our approach on multiple public datasets and showcase the improved performance compared to other neural network-based approaches. In addition, we discuss the limitations and possible further research of our approach for improving the applicability of neural networks to tabular data.\\
		\href{https://es-cloud.cs.uni-tuebingen.de/d/8e2ab8c3fdd444e1a135/?p=\%2FInitializationNeuronalNetworksTabularData\&mode=list}{Link}
	\end{abstract}

	\section{Introduction}
	Deep neural networks are currently the state of the art in image classification~\cite{lecun1989backpropagation}, image generation~\cite{radford2015unsupervised}, image-based regression~\cite{shin2019subtask}, text generation~\cite{zhang2022retgen}, text classification~\cite{yao2019graph}, translation~\cite{zhang2023improving}, audio classification~\cite{mcmahan2018listening}, audio to text~cite{dong2021consecutive}, audio generation~\cite{lin2021exploiting}, and many more. In the domain of tabular data, neural networks are still behind bagged or boosted tree based approaches, although this form of data is one of the most used data structures. For example financial data~\cite{peng2021survey,borisov2022deep}, patient data~\cite{anand2021watermarking,borisov2022deep}, website databases~\cite{chapman2020dataset}, and any form of administration software for companies and universities uses this data structure. The reasons why decision trees still outperform neural networks are numerous. First, decision trees are very efficient in approximating hyperplane boundaries for decision manifolds, which is common in tabular data. In addition, they are easily interpretable in their structure and there are many post-hoc explainability methods for ensembles of decision trees~\cite{lundberg2018consistent,stojic2019explainable}. This is a major factor, especially for critical data like medical patient records or financial databases. The third factor is the lack of an optimal neural network model for tabular data. The breakthrough in image processing came through convolutions~\cite{lecun1989backpropagation} for example and for language processing, transformers set up a new era~\cite{vaswani2017attention}. Since the model and its initialization forms the error plane it, the major problem of neural networks on tabular data is that they struggle to find a good minimum due to their large set of parameters and their initialization. One example here is that the strict random initialization does not allow the network to consider individually features separately from the rest. Based on the structure of fully connected layers, the network could  consider individually features, but it is prevented from doing so due to a local minimum it gets stuck. This is exactly what we are trying to prevent with our binomial initialization.
	
	In contrast to the fast training, high interpretability, and so far best performance on tabular data of decision trees, neural networks also contain advantages, which could make them the method of choice for the future. The first reason for neural networks is, that in case of superior performance on tabular data, all different data sources could be jointly processed. This means that data like images, audio, text, and tabular data could be jointly handled. The next benefit could be the same as for image, text, and audio classification, which is a huge boost in accuracy. In addition, it would reduce the effort for feature engineering, which is currently a key aspect in decision tree-based tabular data learning. Another advantage of neural networks is that they can be trained with batch learning, which only requires a subset of data to be present. This is especially useful for huge datasets. Neural networks are also capable of end-to-end learning of complex tasks like data-efficient domain adaptation~\cite{goodfellow2016deep}, generative modeling~\cite{radford2015unsupervised} and semi-supervised learning~\cite{dai2017good}. 
	
	In this paper, we propose the binomial initialization, which allows the network to see all possible feature combinations separately. Since this is one of the big differences of already published neural network and decision tree-based approaches. In addition, the binomial initialization of neurons can also be used to form feature subsets in the last layer, which enables the joint training of ensembles. Here fore, the same final loss has to be spread over multiple output neurons which try to perform the same task, but the initialization only allows them to see feature subsets. Together with a random  batch gradient masking (Ignoring some batch entries per output neuron), this is similar to subspace training of decision trees. In short, our contributions to the state of the art are:
	\begin{itemize}
		\item Neural network initialization approach for binary classification on tabular data.
		\item Loss functions to jointly train ensembles of neural networks.
	\end{itemize}
	
	\section{Related work}
	So far there are numerous approaches from different areas for applying neural networks to tabular data as well as many trees based boosting approaches. In the realm of feature selection, the authors try to select only a subset of features which are relevant for the task, which could be classification or regression. In general, forward selection and Lasso regularization are used as techniques to attribute feature importance using the entire training data~\cite{guyon2003introduction}. An alternative to these are global methods, instance wise feature selection refers to picking features individually for each
	input~\cite{chen2018learning}. In \cite{chen2018learning} the authors used explainer models to maximize the information between the selected features and the target variable. The authors from \cite{yoon2018invase} used an actor framework for baseline simulation and used this to optimize the selection. 
	
	The so far dominating methods on tabular data are from the realm of tree-based learning, which is also their main application area. For those methods, the main strength relies in global feature selection based on most statistical information gain~\cite{grabczewski2005feature}. Further improvements to those methods are based on ensemble learning, which either trains trees for a consecutive improvement (Boosting) or by major voting over a set of weak learners (Bagging~\cite{breiman1996bagging}). Here fore, random subsets of features and data entries are used to train trees which are combined afterward~\cite{ho1998random}. XGBoost~\cite{chen2016xgboost}, CatBoost~\cite{prokhorenkova2018catboost}, and LightGBM~\cite{ke2017lightgbm} are the most prominent representatives of the state of the art methods for boosting decision trees. Less prominent as the before mentioned methods are the boosting algorithms AdaBoost~\cite{viola2001rapid}, LPBoost~\cite{demiriz2002linear}, RobustBoost~\cite{freund2009more}, RUSBoost~\cite{seiffert2008rusboost}, and TotalBoost~\cite{warmuth2006totally}.
	
	An alternative to the aforementioned approaches is the integration of neural networks into decision trees. In \cite{humbird2018deep} decision trees are replaced by separated blocks of neural networks, but the approach includes a lot of redundancy and is therefore inefficient for learning and representation. Soft neural decision trees~\cite{wang2017using,kontschieder2015deep} use decision functions, which are differentiable to make the approach trainable by the backpropagation algorithm. In contrast to those differentiable decision functions, decision trees use axis-aligned splits based on the features, which cannot be differentiated. The disadvantage of soft neural decision trees is that they lose automatic feature selection, which often degrades performance. As a possible solution for this disadvantage, a soft
	binning function was proposed in \cite{yang2018deep}. In contrast to this, the authors of \cite{ke2018tabnn} proposed a neural network architecture which explicitly utilizes expressive feature combinations which are transferred from a gradient boosted decision tree. \cite{tanno2019adaptive} proposed a neural network which adaptively grows from small blocks of neurons. SAINT~\cite{somepalli2021saint} introduce inters ample attention, contrastive pre-training, and an improved embedding strategy for tabular data. TabNet~\cite{arik2021tabnet} employs soft feature selection with controllable sparsity and is trained end-to-end.
	
	Our approach in contrast to the state of the art initializes the first layer of a neural network with all possible feature combinations. This sets the neural network initial state close to a minimum, since good feature combinations will increase their strength and others will degrade. This layer can also be used to jointly train ensembles of neural networks by placing it at the end. The two techniques of evaluating feature combinations and training ensembles are the major building blocks of decision trees. In its current state our approach contains the disadvantage with too many possible feature combinations which leads to a large memory consumption. Therefore, we do not see our approach as a new algorithm, beating the state of the art. It is a simple and very effective way to initialize neural networks, which could improve all the other approaches. The disadvantage of the too many possible feature combinations can be handled with pruning or adaptively growing neural networks, but this is out of the scope of this paper and will be the focus of further research.

	\section{Method}
	\begin{figure}
		\centering
		\includegraphics[width=0.5\columnwidth]{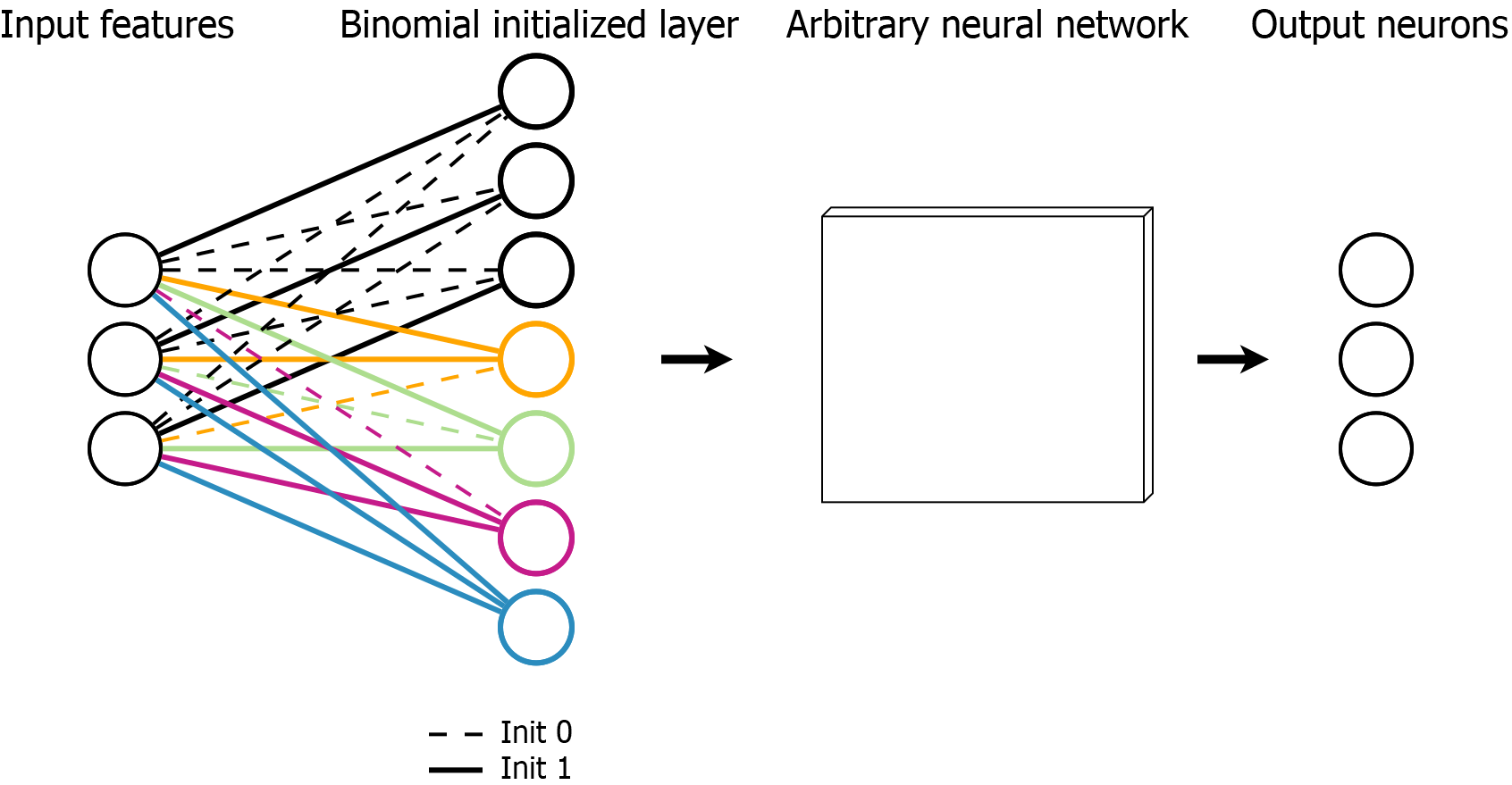} 
		\caption{Structure and placement of the binomial initialized layer in a neural network. The dashed lines are initialized with zero and the solid lines with one. This means our binomial initialized layer is a fully connected layer~\cite{rosenblatt1958perceptron} with a binary initialization, where a combination of features is initialized by one and the remaining connections of this neuron are set to zero. After the binomial initialized layer, which is placed as first layer, an arbitrary neural network can be placed. Based on the initialization, the neural network sees all feature combinations as a starting point. In case of too many features, the amount of neurons can be reduced via a random selection of feature combinations.}
		\label{fig:binoinit}
	\end{figure}
	
	Figure~\ref{fig:binoinit} shows the placement of the binomial initialized layer to enable the neuronal network to see all feature combinations as a starting point. The idea for this initialization steams from random forests. In random forest training, the features are evaluated separately and against each other to compute the information gain. In neural networks, usually the Xavier initialization is used~\cite{glorot2010understanding}, which normalizes the random initialization based on the inputs and outputs of the neuron. For tabular data, this sets the minima to a location which is not explainable and hinders the network to see feature combinations, since local minima force the gradient to oscillate. Our idea is to set the initialization to all feature combinations and let the network select what it needs. From a human perspective, this initialization is totally logical and explainable, since we would not do it differently if we had to select the best features and feature combinations by hand. In Figure~\ref{binoinit} the binomial initialized layer is drawn in color and the solid lines are initialized by one and the dashed lines are set to zero.
	
	\begin{algorithm}[tb]
		\caption{Initialization of all possible combinations of features. Per neuron, all features for the current combination are set to 1 and all the others are set to 0.}
		\label{alg:INITalgorithm}
		\textbf{Input}: weights, inputs\\
		\textbf{Note}: This algorithm expects that the amount of neurons is equal to the possible combinations
		\begin{algorithmic}[1]
			\STATE $weights=0$
			\STATE $AllFeatCombinations=binom(inputs, 1~to~inputs)$ // All feature combinations
			\FORALL{FeatCombinations in AllFeatCombinations}
			\FORALL{Feature in FeatCombinations}
			\STATE $weights[Feature]=1$
			\ENDFOR
			\ENDFOR
		\end{algorithmic}
	\end{algorithm}
	
	\begin{algorithm}[tb]
		\caption{Initialization of random combinations of features. Per neuron, all features for the current combination are set to 1 and all the others are set to 0.}
		\label{alg:INITalgorithmRND}
		\textbf{Input}: weights, inputs, neurons, rnd\\
		\textbf{Note}: This algorithm is used if the possible feature combinations are too many
		\begin{algorithmic}[1]
			\STATE $weights=0$
			\STATE $AllFeatCombinations=binom(inputs, 1~to~inputs)$ // All feature combinations
			\FORALL{neurons}
			\STATE $FeatCombinations=AllFeatCombinations[rnd]$
			\FORALL{Feature in FeatCombinations}
			\STATE $weights[Feature]=1$
			\ENDFOR
			\ENDFOR
		\end{algorithmic}
	\end{algorithm}
	
	Algorithm~\ref{alg:INITalgorithm} and \ref{alg:INITalgorithmRND} are for the binomial initialization. The first algorithm initializes all possible feature combinations and the second algorithm, randomly selects feature combinations, which are initialized. If the feature amount in the dataset is small and all feature combinations can be represented as a neuron, the first algorithm is used. In case of too many features, we used the second algorithm and only evaluated a small subset of possible features. We denote Algorithm~\ref{alg:INITalgorithm} as "Proposed" and Algorithm~\ref{alg:INITalgorithm} as "Prop. RND" in our evaluation.
	
	\begin{figure}[t]
		\centering
		\includegraphics[width=0.95\columnwidth]{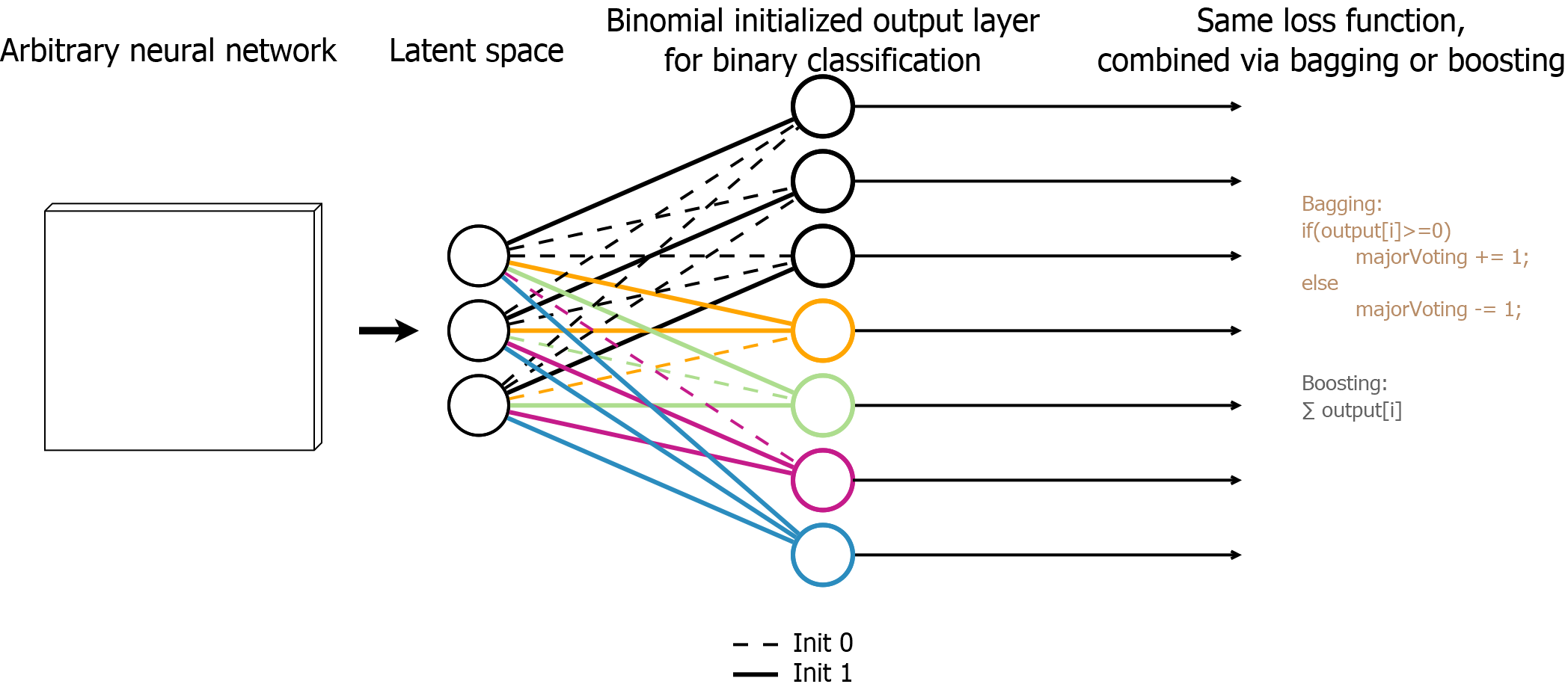}
		\caption{For joint ensemble training, the binomial initialized layer is placed at the end of the entire neural network. Additionally, each neuron of the binomial initialized layer has the same loss function which can be combined using boosting or bagging, as shown on the right side. In addition to the copied loss function, entries in a batch can be masked to further diverge the gradients of each output neuron of the binomial initialized layer.}
		\label{fig:jointens}
	\end{figure}
	
	Figure~\ref{fig:jointens} shows the usage of the proposed binomial initialized layer for joint ensemble training. Here the last layer is a binomial initialized layer with random feature selection since the latent space is usually too large to represent all possible combinations. For the joint ensemble training, we also modified the hinge binary classification loss as well as the soft max loss. 
	
	\begin{equation}
		Hinge(out_i) = \begin{dcases*}
			$-sc*GT$ & $1-GT*out_i > 0$\\
			0 & else
		\end{dcases*}
		\label{eq:hinge}
	\end{equation}
	
	Equation~\ref{eq:hinge} shows the modified hinge loss, $GT$ is the ground truth and either one or minus one, $sc$ is the scale factor, and $out_i$ is the output of the ith neuron in the binomial initialized layer. In addition, the scale factor $sc$ can be used in combination with a random batch instance selection. This means, that only a subset of samples in a batch are used per neuron and the scale factor $sc$ is adapted to the amount of selected entries. We provide the implementation for this also for the binary log loss (Or soft max loss in a multi class setting), but we did not use the random batch instance selection in our evaluation to make the results easier to reproduce.
	
	\begin{equation}
		BinaryLog(out) = \begin{dcases*}
			$GT*sc*(sig( out )-1)$ & $GT > 0$ \\
			$-GT*sc*sig( out )$ & else
		\end{dcases*}
		\label{eq:softmax}
	\end{equation}
	
	Equation~\ref{eq:softmax} shows the modified binary log loss. $sig()$ is the sigmoid function and in a multi class setting, it has to be replaced with the soft max function per output group. This means that the amount of classes forms one group of neurons, which are identically initialized in the binomial initialized layer at the end. The output, similar to the binary case, consists of multiple such groups, where each group has an own feature combination initialization. As for the hinge loss, we also provide the implementation for the random batch instance selection, but we did not use the binary log loss in our evaluation since the hinge loss performed slightly better.
	
	The ensemble itself, as shown in Figure~\ref{fig:jointens} can be trained using bagging and boosting. In our evaluations both performed equally good which is why we only used the boosting in our evaluation. We provide the implementation for both.
	
	\begin{figure}[t]
		\centering
		\includegraphics[width=0.5\columnwidth]{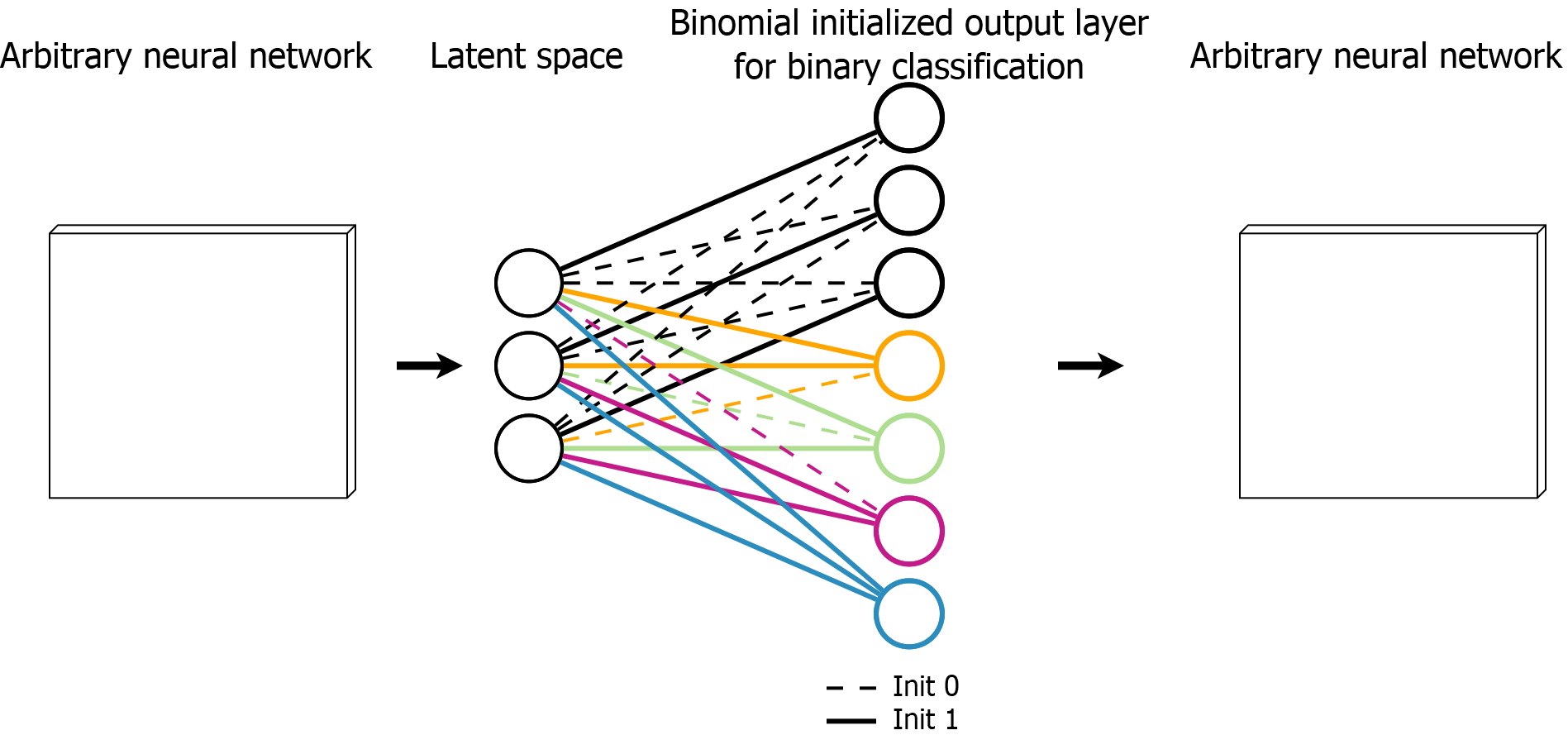}
		\caption{The placement of the binomial initialized layer can be in general everywhere in a neural network. This could be for example useful for autoencoders~\cite{ballard1987modular} or the interconnections in U-Nets~\cite{ronneberger2015u}. For such placements, the random selection of feature combinations should be used since small amount of features already produce immense amounts of possible combinations.}
		\label{fig:inbetween}
	\end{figure}
	
	Figure~\ref{fig:inbetween} shows an alternative usage of our layer. The two DNN blocks could be for example from an auto encoder, where our binomial initialized layer is used to provide different latent feature combinations, but there are numerous other possible applications. One alternative is the usage of U-Nets, where the interconnections between the decoder and encoder could be differently spread in the network. It would also be possible to use it in inception nets instead of the concatenation and form a hierarchical combination of different subnetworks. While there are numerous application areas, we only focused on two tiny and small networks to evaluate the effectiveness of the proposed binomial initialized layer, since all the other applications would break the page limit of this conference.
	
	\begin{figure}[t]
		\centering
		\includegraphics[width=0.51\columnwidth]{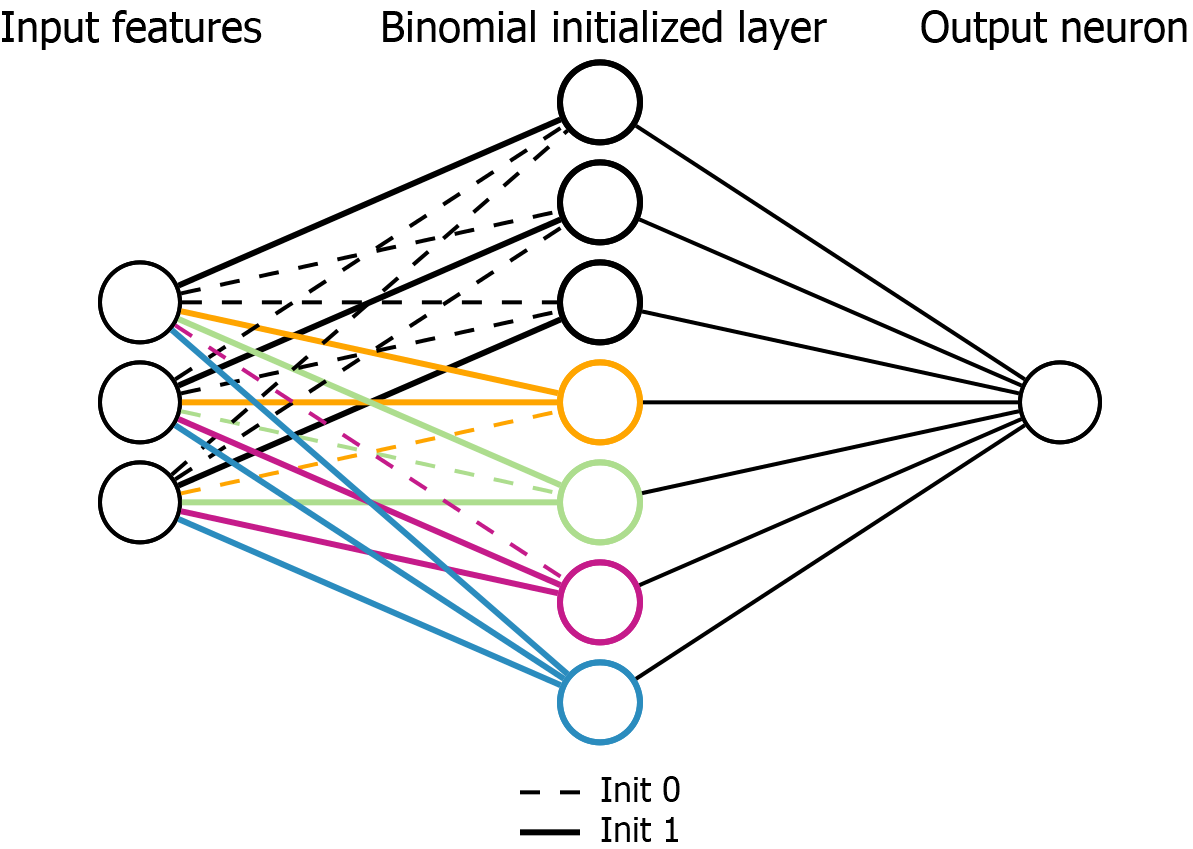}
		\includegraphics[width=0.51\columnwidth]{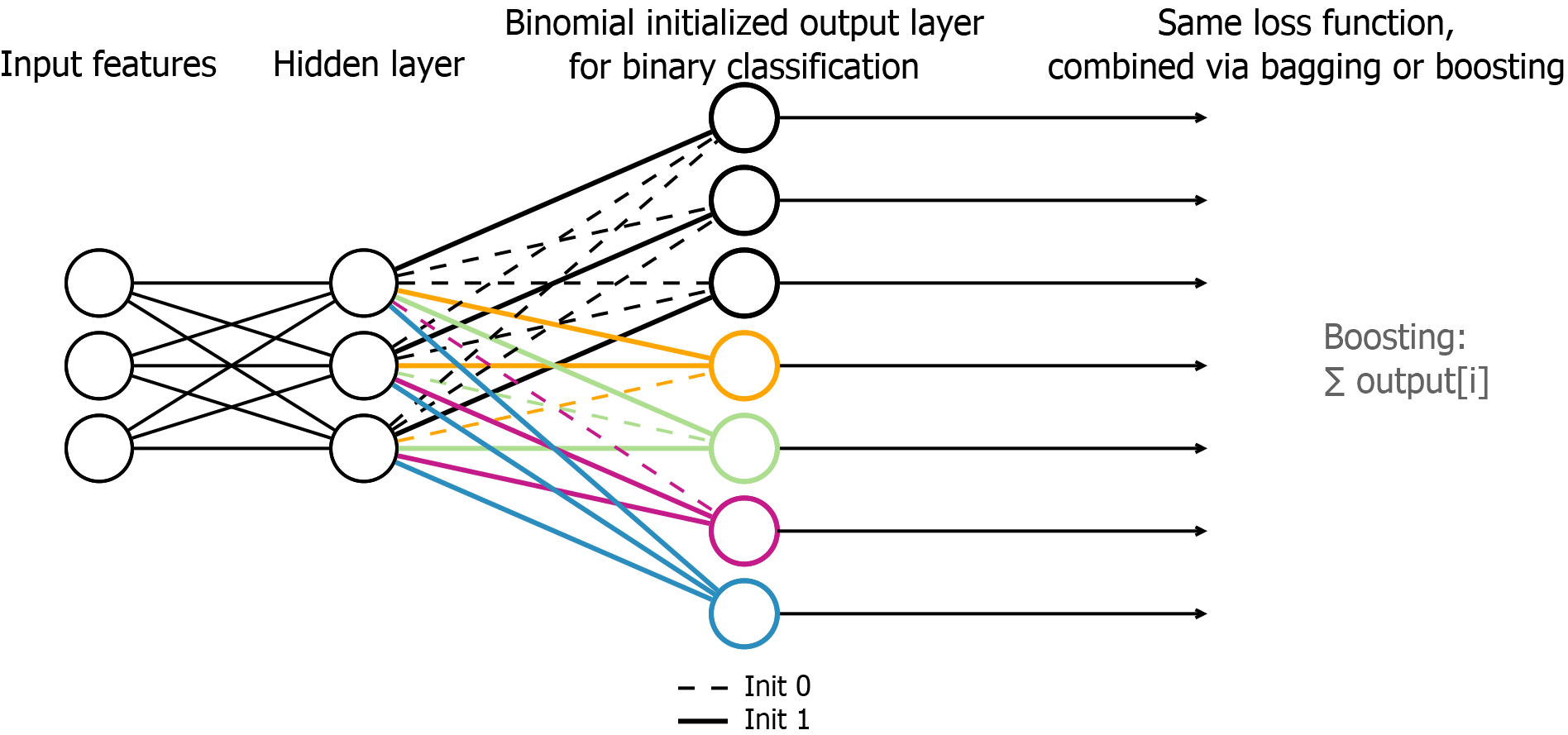}
		\caption{The used architectures in our evaluation. The first network (Top) consists of the binomial initialized layer and one output neuron, since we used the hinge loss~\cite{gentile1998linear}. It is denoted as "Proposed" in our evaluation. The second network is our jointly trained ensemble, which consists of one hidden layer, the binomial initialized layer as output, and the modified hinge loss with boosting. This network is denoted as "Prop. ENS" in our evaluation.}
		\label{fig:usedArch}
	\end{figure}
	
	Figure~\ref{fig:usedArch} shows the used architectures in our evaluation. All networks are small neural networks with one hidden layer. If the first layer is the binomial initialized layer, we denote our network with "Proposed" or "Prop. RND" which is dependent on the amount of features. For the RND initialization, we always used 20k neurons. For the ensemble (Bottom visualization in Figure~\ref{fig:usedArch}) we always set the hidden layer to 256 neurons and the last layer are 1024 neurons which are linearly initialized based on the possible feature combinations. This means each neuron of the 256 hidden neurons is connected separately to one of the first 256 output neurons, and the remaining 768 are combinations of two hidden neurons. This can be done with a random initialization too, but we selected the linear initialization since it is easier to reproduce.

	\section{Datasets}
	In this section, we describe all the used public datasets. We focused only on the binary classification tasks, which is already challenging for modern neural network based approaches. We selected unbalanced datasets like playground-series-s3e3 or playground-series-s3e4 as well as datasets with many features ending up in exceeding the amount of memory we have to also showcase the limitations of our approach. For those datasets we use our normal initialization, random feature combinations as well as our proposed ensemble technique and show the impact on the results in the evaluation section.
	
	\subsection{Adult}
	We downloaded the dataset from 
	\href{http://archive.ics.uci.edu/ml/datasets/adult}{Link} and it consists of 32k entries of training data and 16k entries of testing data. In total, it has 14 features, which sums up to 16k possible feature combinations for our binomial initialization layer. It contains categorical and numerical entries with missing data. The task in this data set is to predict whether income exceeds \$50K per year or not based on census data. It is among the most popular tabular datasets used in modern tabular data classification tasks~\cite{UCIMachine}.
	
	\subsection{HELOC}
	We downloaded the dataset from \href{https://www.kaggle.com/datasets/averkiyoliabev/home-equity-line-of-creditheloc}{Link} and it consists of 10k entries. We made a random train test split and selected 7.8k entries for training and 2.6k for testing. It has 23 features, which sums up to 8.3M possible feature combinations for our binomial initialization layer. The Home Equity Line of Credit (HELOC) dataset is provided by FICO~~\cite{FICO} and is an anonymized dataset from real homeowners. The task here is to use the information about the applicant to predict whether they will repay their HELOC within 2 years. The dataset contains categorical and numerical entries, and the target variable to predict is a binary variable called RiskPerformance.
	
	\subsection{HIGGS}
	We downloaded the dataset from \href{https://archive.ics.uci.edu/ml/datasets/HIGGS}{Link} and it consists of 11M entries. We made a random train test split and selected 8.25M entries for training and 2.75M for testing. It has 28 features which sums up to 268M possible feature combinations for our binomial initialization layer, and it is, therefore, the first dataset which exceed our memory capabilities on our GPU. This dataset was produced by a Monte Carlo simulation, and the first 21 features are kinematic properties measured by the particle detectors. The remaining 7 features are functions of the first 21 features, which are derived by physicists. The task is to distinguish between signals with Higgs bosons (HIGGS) and a background process, and it is therefore also a binary classification dataset. It only contains numerical data without any categorical variable.
	
	\subsection{playground-series: s3e2, s3e3, s3e4, s3e7, s3e10}
	These five datasets are generated from deep learning models trained on different datasets. They contain special challenges like unbalanced data, many features or low amounts of data, and it is ensured that the feature distributions are close to the feature distribution in the original dataset. The datasets are downloaded from \href{https://www.kaggle.com}{Link}  and we used a fixed train test split for each dataset since the test set is not publicly available. For the train test splits we used a random selection ending up in s3e2: 11.4k training, 3.8k testing; s3e3: 1258 training, 419 testing; s3e4: 164k training, 54k testing; s3e7: 31k training, 10k testing; and s3e10: 88k training, 29k testing. For model training, the authors used the following datasets s3e2: Stroke Prediction Dataset; s3e3: Employee Attrition; s3e4: Credit Card Fraud Detection; s3e7: Reservation Cancellation Prediction; and s3e10: Pulsar Classification. The amount of features is s3e2: 10, binomial feature combinations are 1k; s3e3: 33, binomial feature combinations are 8B; s3e4: 30, binomial feature combinations are 1B; s3e7: 17, binomial feature combinations are 131k; and s3e10: 8, binomial feature combinations are 255. Therefore, s3e3 and s3e4 exceed our memory capabilities for a full set of feature combinations for our binomial initialization approach.
	
	\subsection{tabular-playground-series-may-2022}
	This dataset is downloaded from \href{https://www.kaggle.com/competitions/tabular-playground-series-may-2022/data}{Link} and we used a fixed train test split for each dataset since the test set is not publicly available. For the train test splits, we used a random selection with 675k entries for training and 225k for testing. The dataset consists of simulated manufacturing control data and the task is to predict in which state the machine is, which is a binary classification task. In total, it has 31 features which ends up in 2B possible feature combinations and therefore, also exceeds our memory capabilities.
	
	\section{Evaluation}
	The hardware for our training and evaluation is a Windows 10 desktop PC with 64 GB DDR4 RAM, an AMD Ryzen 9 3950X 16-Core Processor @ 3.50 GHz, and an NVIDIA 3090 GPU with 24 GB RAM. For training of CatBoost, LightGBM, and XGBoost we always used the standard parameters of each approach as given by the authors. The evaluation of Bagged, AdaBoost, LPBoost, RobustBoost, RUSBoost, TotalBoost decision trees we used the standard parameters given in Matlab2023a, the only parameter we set manually was the amount of trees for bagged decision trees which we set to 100. For SAINT and TabNet we used the default parameters for binary classification as given by the authors, and we tested them with the raw data and normalized data by mean subtraction and division by two times the standard deviation. The best result is reported in our evaluation (SAINT already integrated the data normalization in the loader function). For our approach we used the ADAM optimizer~\cite{kingma2014adam} with weight decay 0.0005, first momentum 0.9, and second momentum 0.999, a batch size of 1000 and a fixed learning rate of 0.01. The final model is selected by a 20\% validation set which was computed randomly based on the training data, and we trained all models for 30 epochs. During training, we used batch balancing, which means, that the classes are randomly selected, but each class was present in the batch with the same number of samples. All our implementations for the ensemble loss functions as well as the binomial initialized layer are done in dlib~\cite{king2009dlib}.
	
	For our approach, we only used the normalized data (Mean subtraction and division by two times the standard deviation) as input except for s3e10, where we also used the raw data. Categorical features are consecutively numbered, missing values are set to zero and normalized as all the other values.
	
	\definecolor{Gray}{gray}{0.9}
	
	\begin{table*}[t]
		\centering
		\fontsize{8pt}{8pt}\selectfont
		\begin{tabular}{l|c|c|c|c|c|c|c|c|c|c|c|c}
			& \multicolumn{4}{c|}{Adult} & \multicolumn{4}{c}{HELOC} & \multicolumn{4}{c}{HIGGS} \\
			& acc. & prec. & rec. & f1 & acc. & prec. & rec. & f1 & acc. & prec. & rec. & f1\\
			\rowcolor{Gray} Bagged & .854&.737&.597&.660&.716&.734&.647&.688&.754&.766&.775&.770\\
			\rowcolor{Gray} AdaBoost & .849&.731&.572&.642&.716&.719&.676&.697&.685&.696&.720&.708 \\
			\rowcolor{Gray} LPBoost & .833&.693&.531&.601&.688&.696&.630&.661&.530&.530&.999&.692 \\
			\rowcolor{Gray} RobustBoost & .850&.741&.563&.640&.715&.723&.667&.694&.686&.694&.728&.711\\
			\rowcolor{Gray} RUSBoost & .737&.466&.780&.583&.688&.722&.579&.642&.609&.598&.802&.685\\
			\rowcolor{Gray} TotalBoost & .692&.425&.862&.569&.625&.595&.705&.645&.606&.591&.835&.692\\
			\rowcolor{Gray} CatBoost & .872&.774&.648&.705&.720&.731&.667&\textbf{.698}&.756&.764&.782&\textbf{.773} \\
			\rowcolor{Gray} LightGBM &.872&.775&.647&\textbf{.706}&.718&.730&.660&.694&.731&.745&.749&.747 \\
			\rowcolor{Gray} XGBoost & .869&.763&.650&.702&.705&.715&.650&.681&.741&.751&.765&.758 \\ \hline
			SAINT & .814&.685&.395&.501&.706&.713&.656&.684&.733&.736&.774&.754 \\
			TabNet & .848&.744&.546&.630&.712&.706&.694&.700&.745&.780&.723&.750 \\
			MLP 256 & .781&.523&.864&.651&.679&.646&.745&.692&.714&.751&.688&.718 \\
			MLP 100k & .779&.519&.868&.650&.716&.728&.659&.692&.714&.744&.701&.722 \\ \hline
			Proposed & .790&.535&.856&.658&.714*&.693*&.733*&\textbf{.713*}&.715*&.722*&.754*&.737* \\	
			Prop. RND & .795&.543&.840&.659&.718&.724&.672&.697&.721&.735&.741&.738 \\
			Prop. ENS & .798&.548&.833&\textbf{.661}&.705&.678&.742&.708&.742&.747&.775&\textbf{.761} \\
			& \multicolumn{4}{c|}{s3e2} & \multicolumn{4}{c}{s3e3} & \multicolumn{4}{c}{s3e4} \\
			\rowcolor{Gray} Bagged & .959&.566&.106&.179&.916&.750&.081&.146&.997&NaN&.000&NaN \\
			\rowcolor{Gray} AdaBoost & .958&.625&.031&.059&.928&.684&.351&.464&.997&.333&.007&.015 \\
			\rowcolor{Gray} LPBoost & .948&.259&.132&.175&.914&.517&.405&.454&.251&.001&.595&.003 \\
			\rowcolor{Gray} RobustBoost & .958&NaN&.000&NaN&.921&.590&.351&.440&.997&NaN&.000&NaN \\
			\rowcolor{Gray} RUSBoost & .780&.140&.836&.240&.761&.248&.837&.382&.899&.010&.468&.021 \\
			\rowcolor{Gray} TotalBoost & .885&.217&.672&\textbf{.328}&.646&.182&.864&.301&.971&.011&.134&.021\\
			\rowcolor{Gray} CatBoost & .959&.542&.119&.195&.930&1.0&.216&.355&.997&.750&.023&\textbf{.046} \\
			\rowcolor{Gray} LightGBM & .957&.400&.050&.089&.923&.727&.216&.333&.996&.034&.023&.028\\
			\rowcolor{Gray} XGBoost &.957&.441&.119&.188&.937&.789&.405&\textbf{.535}&.997&.600&.023&.045 \\ \hline
			SAINT & .957&.250&.012&.023&.911&NaN&.000&NaN&.997&NaN&.000&NaN \\
			TabNet & .956&.320&.050&.086&.904&.200&.027&.047&.997&NaN&.000&NaN \\
			MLP 256 & .780&.139&.830&.239&.875&.285&.270&.277&.970&.019&.238&.036 \\
			MLP 100k & .815&.142&.685&.236&.880&.290&.243&.264&.983&.020&.126&.035 \\ \hline
			Proposed & .810&.147&.748&\textbf{.246}&.863*&.300*&.405*&.344*&.974*&.021*&.222*&.038* \\ 
			Prop. RND & .790&.141&.798&.240&.883&.333&.324&.328&.980&.026&.214&.047 \\
			Prop. ENS & .781&.135&.792&.231&.868&.312&.405&\textbf{.352}&.994&.060&.111&\textbf{.078} \\		
			& \multicolumn{4}{c|}{s3e7} & \multicolumn{4}{c}{s3e10} & \multicolumn{4}{c}{series-may-2022} \\
			\rowcolor{Gray} Bagged & .808&.767&.731&.749&.991&.969&.937&.953&.826&.831&.805&.818 \\
			\rowcolor{Gray} AdaBoost & .783&.725&.719&.722&.990&.964&.934&.949&.659&.654&.632&.643 \\
			\rowcolor{Gray} LPBoost & .635&.524&.718&.606&.982&.879&.946&.912&.503&.479&.259&.336\\
			\rowcolor{Gray} RobustBoost & .783&.725&.716&.721&.991&.971&.935&.952&.658&.649&.644&.647\\
			\rowcolor{Gray} RUSBoost & .719&.709&.478&.571&.982&.872&.953&.911&.574&.545&.751&.632 \\
			\rowcolor{Gray} TotalBoost & .609&NaN&.000&NaN&.862&.403&.954&.566&.603&.591&.595&.593 \\
			\rowcolor{Gray} CatBoost & .820&.781&.750&.765&.991&.967&.940&.953&.880&.881&.870&\textbf{.875} \\
			\rowcolor{Gray} LightGBM & .816&.775&.747&.761&.991&.969&.940&.954&.832&.836&.814&.825 \\
			\rowcolor{Gray} XGBoost & .819&.777&.754&\textbf{.766}&.991&.970&.940&\textbf{.955}&.859&.860&.847&.853 \\ \hline
			SAINT & .726&.680&.565&.617&.984&.886&.953&.918&.514&NaN&.000&NaN \\
			TabNet & .798&.757&.713&.734&.990&.970&.924&.946&.888&.905&.860&\textbf{.882} \\
			MLP 256 & .790&.714&.771&.742&.986&.900&.961&.930&.850&.843&.849&.846 \\
			MLP 100k & .789&.713&.770&.740&.989&.972&.910&.940&.850&.839&.854&.847 \\ \hline
			Proposed & .794&.714&.789&.750&.989&.940&.943&.941&.849*&.826*&.874*&.849* \\ 
			Prop. RND & .792&.713&.783&.746&.990&.964&.930&\textbf{.947}&.852&.833&.868&.850 \\
			Prop. ENS & .798&.723&.784&\textbf{.752}&.989&.943&.944&.943&.876&.867&.880&.874
		\end{tabular}
		\caption{Shows the results in comparison to SOTA decision tree based approaches (Gray background color) as well as to SOTA neural network based approaches (SAINT and TabNet) and the baseline (MLP). Each result is evaluated five times and the mean value is reported. * indicates, that the amount of feature combinations was too large which is why we set the amount of neurons to 100k, and it has therefore the same size as the MLP 100k. MLP 256 with 256 hidden neurons was chosen, since the Prop. ENS also uses 256 hidden neurons. The best f1 scores for the decision tree based approaches and the neural network based approaches are bold. \textit{acc. is accuracy, prec. is precision, rec. is recall, and f1 is the f1 score.}}
		\label{tbl:results}
	\end{table*}

	Table~\ref{tbl:results} shows the results of our proposed approaches in comparison to SOTA neural network based approaches SAINT and TabNet, the baseline MLP, and SOTA decision tree approaches. Overall, the decision tree based approaches still outperform neural networks, and it has to be mentioned, that we did not use any hyperparameter tuning nor data augmentation for any approach. This means, that all the decision tree based approaches, SAINT, TabNet, the MLP, and our approach can perform much better but for a fair comparison we think this is the best scenario. We also do not want to claim that our approach is the new SOTA, all we want to show, is that the simple binomial initialization of fully connected layers is an effective approach on tabular data. It also opens a variety of possible integrations into deep neural networks, as it was shown for our jointly trained ensemble (Prop. ENS). For "Proposed", where the binomial layer is in the first stage, some results are marked with a *. Those results are computed with 100k neurons only, which means that only the first 100k possible feature combinations are initialized. This was done since all the possible feature combinations are too many for our GPU and our PC RAM. We also evaluated an MLP with 100k neurons as a direct comparison. For the heavily unbalanced dataset s3e4, the main impact of our approach steams from the batch balancing, so the comparison to SAINT and TabNet is somehow unfair, but we trained the MLPs also with batch balancing and here the improvement is directly visible. In general, there are two important things which are visible in the results in Table~\ref{tbl:results}. First, all proposed approaches improve over the MLPs, especially the ensemble (Prop. ENS), which has 256 neurons in the first layer, followed by 1024 binomial initialized neurons forming the multitude of classifiers. This can be directly compared with the MLP with 256 hidden neurons, and it consequently improves over the baseline. The second interesting result is that there is not one superior approach for tabular data in our experiments. This is true for the decision tree based approaches as well as for the neural network based approaches. Sometimes, it is important to use all the different feature combinations (Proposed is best) and for other datasets, ensembles are much better (Prop. ENS is best). For s310, it is also interesting to see, that the 20k neurons which are initialized with random feature combinations outperforms the "Proposed" approach which had all 255 possible combinations initialized. This  means that redundancy can have a positive effect on the result.
	
	For all the decision tree based approaches, CatBoost and XGBoost seam to be the superior approaches in our collection of datasets. For the dataset s3e2 this is not the case, here TotalBoost outperforms all other approaches by a large margin, which indicates, that there is still no superior approach based on decision trees for tabular data. This is also the case for neural networks. While our proposed approaches outperform the other methods except for "series-may-2022", where TabNet is the best neural network based method, this does still not mean, that this is for all datasets the case. It has also to be noted, that our ensemble (Prop. ENS) is an ensemble of standard neural networks with the Xavier initialization. To use both approaches in one neural network, so the ensemble and the feature combinations in the beginning, three layers have to be used (First binomial, second fully, last binomial together with the ensemble hinge loss), which also work very well in our experiments, but due to the page limit we omitted those networks from our evaluation. As can be seen in Figure~\ref{fig:usedArch} we only used very basic neural network architectures to showcase the performance of the binomial initialized layer. TabNet and SAINT are more advanced networks, which could also benefit from our approach.

	\section{Limitations}
	The most obvious limitation of our approach is the exponential growth of the binomial initialized layer. For the dataset "tabular-playground-series-may-2022" with 31 features, the size grows to over 2 billion neurons, which does not fit on any modern GPU based on the RAM capabilities. This limitation can be handled with the proposed random selection of feature combinations, but this approach is just the most obvious and trivial solution. Better solutions would be growing and shrinking networks with Bayesian optimization~\cite{snoek2012practical} capable of improving the network and also exploring the possible feature combination space. An alternative solution would be evolutionary algorithms. Here, multiple nets with feature combinations will be grown, and the strongest will survive. Another limitation of our approach is the multi-class problem, in which it does not perform as well as in the binary case. Here, it would be necessary to form the multi-class classification with binary classifications in an inception-style network. While these limitations are tough in the case of real-world applications, our approach also delivers a multitude of applications in modern deep neural networks like inception nets~\cite{szegedy2017inception}, autoencoders~\cite{ballard1987modular}, or any other deep learning architecture.
	
	\section{Conclusion}
	In this paper, we show that binomial initialization outperforms or performs equal to SOTA neural network-based approaches for binary tabular data classification under the same conditions. We did not perform a hyperparameter search for any approach, which makes it a fair comparison in our opinion. In addition, we gave many possible further research options that will advance the proposed solution, to which we will now also devote ourselves. The binomial layer itself can already be integrated into many deep neural network architectures like autoencoders, inception networks, U-Nets, generative adversarial networks, etc. and could also help in the explanation since the starting point of the network is simple, explainable, and fixed. We also discussed the current limitations of our approach as well as possible solutions. In our opinion, adaptively growing and shrinking networks are the most promising solution, not only for our approach but for the machine learning community in general. This is due to the success of architecture search, which has to be learned by future networks inherently, and in our opinion, adaption is one important aspect of intelligence. Alternatives already known to overcome the limitations of our approach are the prominent pruning of neural networks together with iterative fitting, and the selection of the best initial selection of features, similar to architecture search as we have used it so far in science. Overall, we proposed a simple solution that already improves the performance of neural networks on tabular data, and we hope that this approach will motivate other researchers to advance this approach further.

	\bibliographystyle{plain}
	\bibliography{template}

\end{document}